\documentclass[10pt, a4paper]{article}

\usepackage[]{lrec-coling2024} % this is the new style
\usepackage{amsmath}
\usepackage{tikz-dependency}
\usepackage{multirow}
\title{Polish-ASTE: Aspect-Sentiment Triplet Extraction Datasets for~Polish}

\name{Marta Lango$^*$, Borys Naglik$^2$, Mateusz Lango$^{1,3}$, Iwo Naglik$^{1,4}$} 

\address{$^1$ Poznan University of Technology, Faculty of Computing and Telecommunications\\
$^2$ Adam Mickiewicz University, Faculty of Mathematics and Computer Science \\
$^3$Charles University, Faculty of Mathematics and Physics, Institute of Formal and Applied Linguistics \\
$^4$ deepsense.ai, (research@deepsense.ai)
$^*$ independent researcher
}

\abstract{
Aspect-Sentiment Triplet Extraction (ASTE) is one of the most challenging and complex tasks in sentiment analysis. It concerns the construction of triplets that contain an aspect, its associated sentiment polarity, and an opinion phrase that serves as a rationale for the assigned polarity.
Despite the growing popularity of the task and the many machine learning methods being proposed to address it, the number of datasets for ASTE is very limited. % and includes only English datasets. 
In particular, no dataset is available for any of the Slavic languages.
In this paper, we present two new datasets for ASTE containing customer opinions about hotels and purchased products expressed in Polish.
We also perform experiments with two ASTE techniques combined with two large language models for Polish to investigate their performance and the difficulty of the assembled datasets.
The new datasets are available under a permissive licence and have the same file format as the English datasets, facilitating their use in future research.
 \\ \newline  }
\newcommand*{\tabindent}{ \hspace{3mm}}
\begin{document}

\maketitleabstract

\section{Introduction}

\begin{figure*}
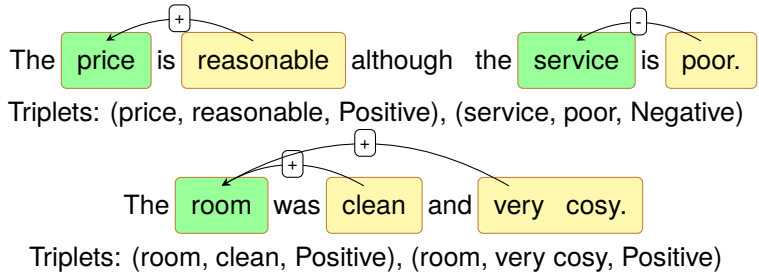

\centering
\begin{dependency}[arc edge, arc angle=30]
\begin{deptext}[column sep=.1cm]
The \& price \& is \& reasonable \& although \& the \& service \& is \& poor.\\
\end{deptext}
\depedge{4}{2}{+}
\depedge{9}{7}{-}
\wordgroup[group style={fill=green!40, draw=brown, inner sep=.6ex}]{1}{2}{2}{a0}
\wordgroup[group style={fill=green!40, draw=brown, inner sep=.6ex}]{1}{7}{7}{a0}
\wordgroup[group style={fill=yellow!40, draw=brown, inner sep=.6ex}]{1}{4}{4}{a0}
\wordgroup[group style={fill=yellow!40, draw=brown, inner sep=.6ex}]{1}{9}{9}{a0}
\end{dependency}\\
Triplets: (price, reasonable, Positive), (service, poor, Negative)\\
\begin{dependency}[arc edge, arc angle=30]
\begin{deptext}[column sep=.1cm]
The  \& room \&  was \&  clean \&  and \& very \&  cosy.\\
\end{deptext}
\depedge{4}{2}{+}
\depedge{6}{2}{+}
\wordgroup[group style={fill=green!40, draw=brown, inner sep=.6ex}]{1}{2}{2}{a0}
\wordgroup[group style={fill=yellow!40, draw=brown, inner sep=.6ex}]{1}{4}{4}{a0}
\wordgroup[group style={fill=yellow!40, draw=brown, inner sep=.6ex}]{1}{6}{7}{a0}
\end{dependency}\\
Triplets: (room, clean, Positive), (room, very cosy, Positive)
    \caption{ASTE triplets extracted from two example sentences. %~\cite{naglik23}. 
    Opinion and aspect phrases are highlighted in yellow and green, respectively. The +/- sign denote positive/negative sentiment.}
    \label{fig:example}
\end{figure*}

Aspect Sentiment Triplet Extraction (ASTE) is a recently proposed sentiment analysis (SA) task~\cite{peng2020knowing} that 
involves the extraction of triplets comprising:
\begin{itemize}
    \item     aspect phrase - a text span that represents a particular feature or attribute of the item, for which an opinion is being expressed
    \item sentiment polarity - often categorised as positive, negative or neutral and refers to the emotional tone being expressed regarding the given aspect.
    \item opinion phrase - a text span that explicitly conveys the sentiment towards the aspect. 
\end{itemize}
Two examples of sentences with extracted ASTE triplets %can be found in Figure~\ref{fig:example}.
are given below and in Figure~\ref{fig:example}.
\begin{quote}
Cena jest rozsądna, ale obsługa słaba.
\emph{The price is reasonable, but the service is poor.}\\
Triplets: (Cena$_{\text{price}}$, rozsądna$_{\text{reasonable}}$, Positive), (obsługa$_{\text{service}}$, słaba$_{\text{poor}}$, Negative)

Pokój był czysty i bardzo przytulny.\\
\emph{The room was clean and very cosy.} \\
Triplets: (Pokój$_{\text{room}}$, czysty$_{\text{clean}}$, Positive), (Pokój$_{\text{room}}$, bardzo przytulny$_{\text{very cosy}}$, Positive)
\end{quote}
Note that multi-word phrases and one-to-many links are possible.

Since the ASTE triples provide the comprehensive "What, How and Why" information regarding the sentiment~\cite{peng2020knowing}, the task quickly gained significant research attention.
Many machine learning techniques have been proposed, including GTS~\cite{wu-etal-2020-grid}, JET~\cite{jet}, Span-based approach~\cite{span-level},  PBF~\cite{more-fine-grained}, GAS~\cite{T5}, a two-stage approach \cite{First_Target_and_Opinion_then_Polarity}, BMRC~\cite{liu-etal-2022-robustly}, SBC \cite{sbc} and EPISA~\cite{naglik23}.
However, the experimental evaluation of these models is limited to English only, since datasets for other languages are not available.

In this paper, we introduce two novel datasets for ASTE that contain customer opinions on hotels and purchased products, all expressed in Polish. Furthermore, we conduct experiments using two ASTE techniques in conjunction with two state-of-the-art large language models designed for Polish to assess the difficulty posed by the newly curated datasets to already existing ASTE methods.

% In this paper, we present two new datasets for ASTE containing customer opinions about hotels and purchased products expressed in Polish.
% We also perform experiments with two ASTE techniques combined with two large language models for Polish to investigate the difficulty of the assembled datasets, at the same time performing the first performance evaluation of these approaches for other language than English.
% The experimental results reveal that the newly proposed datasets are chalanging for the existing methods.   
    
\section{Related works}

There are several datasets constructed for sentiment analysis in Polish.
PolEmo 2.0~\cite{kocon-etal-2019-multi} contains over 57k sentences with annotated sentiment polarity, later extended to AspectEmo~\cite{11321849} with sentiment assigned to each token of the sentence. Even more fine-grained annotation can be found in the Polish Sentiment Treebank\footnote{http://zil.ipipan.waw.pl/TreebankWydzwieku}, used in the PolEval-2017 challenge~\cite{wawer2017results}, which is a dependency treebank with sentiment annotations for each subphrase of the sentence.

Short informal texts are at the focus of TwitterEmo corpus~\cite{10.1007/978-3-031-36021-3_20} which provides annotation for detecting sarcasm, sentiment and basic emotions.
\citet{ptaszynski2019results} collected dataset for cyberbullying detection.
Similarly, HateSpeech corpus\footnote{\url{http://zil.ipipan.waw.pl/HateSpeech}}\cite{troszynski2017czy}  comprises online posts with offensive content.

Allegro Reviews dataset~\cite{rybak-etal-2020-klej} contains customer reviews with 1-5 ratings from a popular Polish e-commerce marketplace.
Finally, Wroclaw Corpus of Consumer Reviews Sentiment (WCCRS,~\citealt{Kocon2019}) is a multi-domain dataset of Polish textual reviews, available on a permissive license that allows its adaptation and transformation.

None of the aforementioned Polish datasets is suitable for training machine learning models for ASTE, but naturally such datasets do exist for English.
The datasets originate from SemEval shared tasks~\cite{pontiki-etal-2014-semeval,pontiki-etal-2015-semeval,pontiki-etal-2016-semeval}, for which the following corpora were constructed: 14lab (opinions about laptops), 14res, 15res and 16res (restaurant reviews).
 \citet{fan-etal-2019-target} extended these corpora by adding opinion term annotation and \citet{peng2020knowing} %and~\citet{wu-etal-2020-grid} 
 finally compiled them for ASTE.

It is important to mention that there are other related tasks like structured SA or fine-grained SA, which have datasets constructed for different languages  e.g.~for {B}asque and {C}atalan~\cite{barnes-etal-2018-multibooked}, German \cite{klinger-cimiano-2014-usage}, Czech~\cite{steinberger-etal-2014-aspect} or Hindi~\cite{akhtar-etal-2016-aspect}.
Some of these tasks even involve the extraction of some triplets from texts, but they contain opinion holders, aspect categories, or others.

\section{New ASTE Datasets for Polish }
\subsection{Construction of datasets}
The customer reviews for the new datasets were taken from the training part of Wroclaw Corpus of Consumer Reviews Sentiment (WCCRS)~\citelanguageresource{WCCRS}, which contains online customer reviews from four domains: medicine, hotels, products and students' opinions about lectures. 
%Only reviews from the training part of WCCRS were included.
The sentiment polarity in WCCRS is provided at the sentence and document level, but not at the aspect level. Therefore, the sentiment annotation from WCCRS was not used during the construction of our ASTE datasets.

We selected two WCCRS domains: hotels (reviews of hotels, originally taken from TripAdvisor) and products (buyers’ opinions on products from Ceneo.pl). 
Following previous work for English \cite{fan-etal-2019-target,peng2020knowing}, we annotated (aspect phrase, opinion phrase, sentiment polarity) triples in each sentence, where aspect/opinion phrases are text spans and the considered sentiment polarities are: positive, negative and neutral.

The datasets were annotated by Polish native speakers using  Doccano annotation platform~\cite{doccano}. Every annotator studied the annotation guidelines prepared by an NLP expert with experience in sentiment analysis. All doubts and questions of the annotators were discussed during the Q\&A sessions.
Quality control was done by reviewing the annotators' output by NLP experts who were allowed to correct them if needed (less than 5\% of examples). 
To measure the inter-annotator agreement, we asked one of our annotators to re-annotate 50 examples from the hotel datasets that had previously been annotated by another annotator. The annotator provided the same annotation for 78\% of the sentences.

A basic summary of the annotation guidelines is provided below.
\begin{itemize}
    \item Aspects should be properties of the whole item/service being described in a review. For instance, in a hotel review the aspects might be “room” or “parking lot” but not “kettle”.
 \item Opinion phrases are the shortest possible phrases that provide a sentiment polarity for an aspect, without changing the intensity of the sentiment polarity.  For example, in the text “very good room” the phrase “very good” is an opinion phrase because a shorter phrase “good” would lose information about the sentiment intensity.
 \item If the review contains an opinion phrase but no aspect, it should not be annotated. 
 \item If the opinion can be assigned to the phrase that denotes the whole described item, it should be treated as an aspect. However, if it is possible to associate the opinion phrase with a more precise aspect, the name of the whole item should be ignored.
 For instance:  "Everything is good" will not be annotated as there is no aspect phrase, "The hotel is good" should yield the (hotel, good, positive) triplet, but "The hotel has good rooms" should be annotated as (rooms, good, positive). 
 \item The aspect phrases should not contain prepositions, even if they determine the noun declension e.g. "W pokoju wszystko było okej" (In the room everything was good) will be annotated as (pokoju/room, okej/good, positive) 
 \item While assigning sentiment polarity, the context and the intention of the opinion giver should be taken into account.
% \item Opinion phrases should be informative even without the context of the sentence.
\end{itemize}

 While annotating WCCRS corpus, we noticed that some of the texts did not appear to be customer reviews, but rather press releases discussing, for example, hotel construction or the tourism industry. Such sentences were flagged by our annotators and excluded from further analysis in this paper. However, we release them along with our annotations for possible future experiments with out-of-domain texts.
 
Our datasets are available under open source licence CC BY-SA 4.0 Deed\footnote{\url{https://creativecommons.org/licenses/by-sa/4.0/}}  on our GitHub repository\footnote{\url{https://anonymous.4open.science/r/Polish-ASTE-Datasets-anonymous/}} with the file format being compatible with English datasets constructed for ASTE~\cite{wu-etal-2020-grid}. 
This will allow easier integration with already existing code bases and hopefully will incentivise comparison of ASTE models on the newly proposed benchmark for Polish.

\begin{table*}[]
%\small 
\begin{tabular}{p{5.5cm}|rrrr|rr|rr}
& \multicolumn{4}{c|}{English}   & \multicolumn{2}{c|}{Polish}    &  \multicolumn{2}{c}{Norm. avg.}             \\
& 14lap & 14res & 15res & 16res & hot.     & prod.   &Eng.&Pol.               \\\hline
Number of sentences                                                                      & 1453 & 2068 & 1075 & 1393 & 590  & 511      &  n/a   & n/a            \\\hline
Number of triplets                                                                       & 2349 & 3909 & 1747 & 2247 & 1197 & 851  & 1.69  & 1.85  \\
\tabindent incl. with negative sentiment                                                        & 774  & 754  & 401  & 483  & 541  & 376  & 0.40  & 0.83  \\
\tabindent incl. with neutral sentiment                                                         & 225  & 286  & 61   & 90   & 58   & 54   & 0.10  & 0.10  \\    
\tabindent incl. with positive sentiment                                                          & 1350 & 2869 & 1285 & 1674 & 598  & 421  & 1.18  & 0.92  \\ \hline
Number of aspect phrases                                                             & 2030 & 3392 & 1507 & 1946 & 798  & 693  & 1.46  & 1.35  \\
\tabindent incl. single word aspect phrases                                         & 1292 & 2545 & 1102 & 1427 & 681  & 526  & 1.04  & 1.09  \\
\tabindent incl. multi-word aspect phrases                                           & 738  & 847  & 405  & 519  & 117  & 167  & 0.42  & 0.26  \\\hline
Number of opinion phrases                                                            & 2030 & 3409 & 1620 & 2078 & 1156 & 827  & 1.51  & 1.79  \\
\tabindent incl. single word opinion phrases                                              & 1705 & 3037 & 1421 & 1829 & 412  & 343  & 1.32  & 0.68  \\
\tabindent incl. multi-word opinion phrases                                           & 325  & 372  & 199  & 249  & 744  & 484  & 0.19  & 1.10  \\\hline
Number of one-to-many relation  & 535  & 812  & 307  & 388  & 323  & 150  & 0.33  & 0.42  \\
\tabindent incl. one aspect-to-many opinions                                           & 281  & 443  & 208  & 263  & 289  & 128  & 0.20  & 0.37  \\
\tabindent incl. one opinion-to-many aspects                                           & 254  & 369  & 99   & 125  & 34   & 22   & 0.13  & 0.05  \\\hline
N. of triplets w/ single words spans                                   & 1305 & 2631 & 1140 & 1478 & 385  & 287  & 1.07  & 0.61  \\
N. of triplets w/ a multi-word phrases                       & 1044 & 1278 & 607  & 769  & 812  & 564  & 0.61  & 1.24  \\
\tabindent incl.  with multi-word opinion and single word aspect  & 207  & 302  & 149  & 188  & 649  & 377  & 0.14  & 0.92  \\
\tabindent incl.  with multi-word aspect and single-word opinion  & 684  & 875  & 403  & 513  & 46   & 70   & 0.41  & 0.11  \\\hline
Mean sentence length (words)                                                    & 18.4 & 16.9 & 15.0 & 14.9 & 16.4 & 21.0 & 16.43 & 18.50 \\
Mean length of aspect phrases                        & 1.47 & 1.40 & 1.45 & 1.44 & 1.26 & 1.40 & 1.44  & 1.32  \\
Mean length of opinion phrases                    & 1.25 & 1.16 & 1.19 & 1.19 & 2.97 & 2.22 & 1.20  & 2.62          \\

\hline
\end{tabular}
\caption{Quantitative characteristics of the proposed corpora for Polish (\underline{hot}el, \underline {prod}ucts) and, for comparison. of already existing corpora for English.
Additionally, an average normalized by the size of the corpus (number of sentences) is computed over all corpora for a given language (Norm. avg.). 
}
\label{tab:stats}
\end{table*}
\subsection{Characteristics of new datasets}
The qualitative analysis of the constructed datasets can be found in Table~\ref{tab:stats}. 
For comparison, the same statistics were computed for four English datasets from SemEval competitions. Furthermore, to facilitate comparisons, we computed a weighted average of each statistic value over all datasets for each language, using the size of the corpora (measured by the number of sentences) as the average weight.

The average number of triples in a sentence is slightly higher in the Polish datasets, mostly due to more frequent opinion phrases forming triples with the same aspect.
An important feature of the proposed datasets that makes them more challenging are multi-word opinion phrases, which are almost six times more frequent than in the English datasets.
The average length of an opinion phrase is 2.97 words in the hotels dataset and 2.22 words in the products' dataset, while this value does not exceed 1.25 in any of the English corpora.
Similarly, the number of triples containing solely single word spans in both aspect and opinion phrases is about 1.75 times lower in the Polish datasets.
Such triples are much easier to construct by machine learning methods, making the Polish datasets more challenging for them.

 The distribution of sentiment polarity in the Polish datasets is significantly more balanced between positive and negative classes than in the English counterparts. However, in datasets for both languages, the triplets with Neutral sentiment are quite rare. Note, that the issue of class imbalance is common in sentiment classification~\cite{lango2019tackling}.
\section{Experimental evaluation}

\begin{table*}[t]
\centering
\begin{tabular}{ll|lll|lll}
                         &                     & \multicolumn{3}{c|}{GTS}    & \multicolumn{3}{c}{EPISA}  \\
         Language                &       Dataset              & Precision & Recall & F1    & Precision & Recall & F1    \\\hline
\multirow{4}{*}{English} & 14lap               & 58.54     & 50.65  & 54.30  & 66.98     & 60.55  & 63.56 \\
                         & 14res               & 68.71     & 67.67  & 68.17 & 75.29     & 72.56  & 73.89 \\
                         & 15res               & 60.69     & 60.54  & 60.61 & 66.44     & 64.74  & 65.54 \\
                         & 16res               & 67.39     & 66.73  & 67.06 & 71.12     & 72.45  & 71.77 \\\hline
\multirow{4}{*}{Polish}  & products (TrelBERT) & 45.15     & 40.17  & 41.74 & 50.01     & 43.36  & 46.07 \\
                         & products (HerBERT)  & 40.65     & 33.79  & 35.65 & 48.18     & 44.09  & 45.89 \\
                         & hotels (TrelBERT)   & 42.07     & 37.82  & 39.08 & 49.07     & 41.66  & 44.72 \\
                         & hotels (HerBERT)    & 37.35     & 28.28  & 30.02 & 48.79     & 42.76  & 45.47\\\hline
\end{tabular}
\caption{The experimental results of ASTE task, measured with Precision, Recall and F1 for two methods: GTS and EPISA.
For Polish datasets two variants of these methods are explored, with different backbone masked language models: TrelBERT and HerBERT.
}
\label{t:res}
\end{table*} 

\subsection{Experimental setup}
The aim of experimental evaluation is to assess the difficulty of the newly constructed datasets for Polish.
We selected two deep learning approaches for the comparison:
\begin{itemize}
    \item Grid Tagging Scheme (GTS, \citealp{wu-etal-2020-grid}) is a classical approach for ASTE which builds a classifier predicting a particular word-by-word matrix. Due to the special coding scheme used, the matrix can be converted to aspect-sentiment triplets by a dedicated decoding algorithm.
    \item Exploiting Phrase Interrelations Span-level Approach (EPISA,~\citealp{naglik23}) is a recent technique that first generates all suitable phrases from a given text, and then constructs final triples with a 2-dimensional CRF model that exploits interrelations between the phrases. To the best of our knowledge, this is the state-of-the-art approach for ASTE for English.
\end{itemize}
To construct text representation, the aforementioned models leverage masked language models (MLM) like BERT~\cite{devlin-etal-2019-bert} or DeBERTa~\cite{he2020deberta} which are pre-trained on English texts.
Therefore, in our study we reimplemented them replacing the original MLMs with two established architectures for Polish, namely HerBERT~\citelanguageresource{mroczkowski-etal-2021-herbert} and TrelBERT~\citelanguageresource{szmyd-etal-2023-trelbert}.

Following related works, the models are evaluated using three criteria: precision, recall, and F1-score. An aspect/opinion phrase is deemed accurate only if it perfectly aligns with the gold standard (exact match). A triplet is considered a true positive only if all its parts are correct. All metric values presented were calculated on the respective test set and averaged across ten separate training runs.
60\% of dataset examples were used for training set and 20\% for validation and test set, respectively.
Examples without any annotated triplets  were omitted during training.

\subsection{Results}

The results are presented in Table~\ref{t:res}.
As expected, the more recent EPISA method produced better results in terms of F1 score for all datasets, with an average improvement of 7.5 percentage points over GTS.
For both Polish datasets, the combination of GTS with TrelBERT yielded better results than its HerBERT counterpart. Similarly, the TrelBERT variant achieved the highest F1-score value for EPISA on products corpora.
On the other hand, HerBERT performed slightly better on hotels corpora when paired with EPISA.

The proposed Polish datasets proved to be much more challenging for existing methods than the existing English datasets.
The F1 score averaged over all English datasets is 63.5\% for GTS and 68.7\% for EPISA, while the average performance of the generally better performing TrelBERT variants is 40.5\% and 45.4\% for GTS and EPISA respectively.
The difference in triplet extraction performance between Polish and English is approx. 23 percentage points for both methods, which calls for future research in ASTE for under-resourced languages.

\section{Summary}
In this paper we have presented two datasets for the Aspect Sentiment Triplet Extraction task  for the Polish language.
The new datasets have the same format as commonly used English datasets, which facilitates the comparison and development of machine learning ASTE models for both languages.
The ASTE datasets for Polish are characterised by a higher average number of triplets and a higher frequency of one-to-many relations than their English counterparts. Multi-word opinion phrases are also significantly more common.

The conducted experiments with two ASTE deep learning approaches adopted to Polish by coupling them with two different Polish language models showed that the constructed datasets are challenging for modern machine learning techniques.

\section{Frequently Asked Questions}
\begin{enumerate}
    \item The data used for annotation is a subset of an existing data set - why is it being presented as two datasets?

The research on ASTE is almost exclusively carried out on the English datasets discussed in our paper. These datasets contain texts from only one domain, e.g. opinions about restaurants or laptops. The performance achieved by the current SOTA methods on English is still not fully satisfactory and production-ready to move to a more challenging scenario of multi-domain datasets. As we expected that training ASTE for Polish may be even more challenging, we considered it necessary to produce single-domain datasets as well. However, just as all four English datasets can be merged into one larger dataset, the same is true for the Polish datasets. One can also merge all the datasets in our work to test the multilingual multi-domain scenario.
\item The lower metric values obtained on Polish datasets are interpreted as the datasets being more challenging. Is it possible that the model replication/implementation for Polish has issues? 

Both the GTS and EPISA algorithms have been run by us for Polish and English using original implementation provided by the authors. 
For EPISA we were able to reproduce the exact results of the original work on English, as the original implementation of EPISA was available with fixed random seed.
GTS results differed slightly from the original work on English, but none of the differences were statistically significant.
The Polish datasets have the same input format as the English datasets, and the only change with respect to the original methods is to replace an English MLM with a Polish MLM, so it is unlikely that there are any implementation issues for Polish.

\item Is it possible that the methods used were just optimized for English data, and this causes the lower performance on the Polish data?

Although current SOTA methods based on deep learning do not appear to explicitly model language-dependent features that would favour English over Polish, it is possible that they are inadvertently optimised for English data. For example, published methods tend to perform better on English benchmarks, leading to the promotion of such methods during development, despite the fact that other design choices might work better for other languages.
Developing datasets for languages other than English is the first step in investigating this issue.

\end{enumerate}

\section{Acknowledgements}
The research has been partially supported by 0311/SBAD/0743 PUT University grants.
M. Lango was supported by National Science Centre, Poland (Grant No.~2022/47/D/ST6/01770).
This work used resources of the LINDAT/\hspace{0mm}CLARIAH-CZ Research Infrastructure (Czech Ministry of Education, Youth, and Sports project No.~LM2018101).
For the purpose of Open Access, the author has applied a CC-BY public copyright licence to any Author Accepted Manuscript (AAM) version arising from this submission.

\clearpage
%
%\nocite{*}
\section{Bibliographical References}\label{reference}
%\label{main:ref}

\bibliographystyle{lrec_natbib}
\bibliography{lrec-coling2024-example}

\begin{thebibliography}{3}
\expandafter\ifx\csname natexlab\endcsname\relax\def\natexlab#1{#1}\fi

\bibitem[{Koco{\'n} et~al.(2019)Koco{\'n}, Za{\'s}ko-Zieli{\'n}ska,
  Mi{\l}kowski, Janz, and Piasecki}]{WCCRS}
Koco{\'n}, Jan and Za{\'s}ko-Zieli{\'n}ska, Monika and Mi{\l}kowski, Piotr and
  Janz, Arkadiusz and Piasecki, Maciej. 2019.
\newblock \emph{Wroclaw Corpus of Consumer Reviews Sentiment ({WCCRS})}.
\newblock \href {http://hdl.handle.net/11321/700} {[link]}.

\bibitem[{Mroczkowski et~al.(2021)Mroczkowski, Rybak, Wr{\\'o}blewska, and
  Gawlik}]{mroczkowski-etal-2021-herbert}
Mroczkowski, Robert and Rybak, Piotr and Wr{\\'o}blewska, Alina and Gawlik,
  Ireneusz. 2021.
\newblock \emph{{H}er{BERT}: Efficiently Pretrained Transformer-based Language
  Model for {P}olish}.
\newblock Association for Computational Linguistics.
\newblock \href {https://www.aclweb.org/anthology/2021.bsnlp-1.1} {[link]}.

\bibitem[{Szmyd et~al.(2023)Szmyd, Kotyla, Zobni{\'o}w, Falkiewicz, Bartczuk,
  and Zygad{\l}o}]{szmyd-etal-2023-trelbert}
Szmyd, Wojciech and Kotyla, Alicja and Zobni{\'o}w, Micha{\l} and Falkiewicz,
  Piotr and Bartczuk, Jakub and Zygad{\l}o, Artur. 2023.
\newblock \emph{{T}rel{BERT}: A pre-trained encoder for {P}olish {T}witter}.
\newblock Association for Computational Linguistics.
\newblock \href {https://aclanthology.org/2023.bsnlp-1.3} {[link]}.

\end{thebibliography}


\begin{thebibliography}{30}
\expandafter\ifx\csname natexlab\endcsname\relax\def\natexlab#1{#1}\fi

\bibitem[{Akhtar et~al.(2016)Akhtar, Ekbal, and
  Bhattacharyya}]{akhtar-etal-2016-aspect}
Md~Shad Akhtar, Asif Ekbal, and Pushpak Bhattacharyya. 2016.
\newblock \href {https://aclanthology.org/L16-1429} {Aspect based sentiment
  analysis in {H}indi: Resource creation and evaluation}.
\newblock In \emph{Proceedings of the Tenth International Conference on
  Language Resources and Evaluation ({LREC}'16)}, pages 2703--2709,
  Portoro{\v{z}}, Slovenia. European Language Resources Association (ELRA).

\bibitem[{Barnes et~al.(2018)Barnes, Badia, and
  Lambert}]{barnes-etal-2018-multibooked}
Jeremy Barnes, Toni Badia, and Patrik Lambert. 2018.
\newblock \href {https://aclanthology.org/L18-1104} {{M}ulti{B}ooked: A corpus
  of {B}asque and {C}atalan hotel reviews annotated for aspect-level sentiment
  classification}.
\newblock In \emph{Proceedings of the Eleventh International Conference on
  Language Resources and Evaluation ({LREC} 2018)}, Miyazaki, Japan. European
  Language Resources Association (ELRA).

\bibitem[{Bogdanowicz et~al.(2023)Bogdanowicz, Cwynar, Zwierzchowska, Klamra,
  Kiera{\'{s}}, and Kobyli{\'{n}}ski}]{10.1007/978-3-031-36021-3_20}
Stanis{\l}aw Bogdanowicz, Hanna Cwynar, Aleksandra Zwierzchowska, Cezary
  Klamra, Witold Kiera{\'{s}}, and {\L}ukasz Kobyli{\'{n}}ski. 2023.
\newblock Twitteremo: Annotating emotions and sentiment in polish twitter.
\newblock In \emph{Computational Science -- ICCS 2023}, pages 212--220, Cham.
  Springer Nature Switzerland.

\bibitem[{Chen et~al.(2022)Chen, Chen, Sun, and Zhang}]{sbc}
Yuqi Chen, Keming Chen, Xian Sun, and Zequn Zhang. 2022.
\newblock Span-level bidirectional cross-attention framework for aspect
  sentiment triplet extraction.
\newblock In \emph{Proceedings of the 2022 Conference on Empirical Methods in
  Natural Language Processing (EMNLP)}. Association for Computational
  Linguistics.

\bibitem[{Devlin et~al.(2019)Devlin, Chang, Lee, and
  Toutanova}]{devlin-etal-2019-bert}
Jacob Devlin, Ming-Wei Chang, Kenton Lee, and Kristina Toutanova. 2019.
\newblock \href {https://doi.org/10.18653/v1/N19-1423} {{BERT}: Pre-training of
  deep bidirectional transformers for language understanding}.
\newblock In \emph{Proceedings of the 2019 Conference of the North {A}merican
  Chapter of the Association for Computational Linguistics: Human Language
  Technologies, Volume 1 (Long and Short Papers)}, pages 4171--4186,
  Minneapolis, Minnesota. Association for Computational Linguistics.

\bibitem[{Fan et~al.(2019)Fan, Wu, Dai, Huang, and Chen}]{fan-etal-2019-target}
Zhifang Fan, Zhen Wu, Xin-Yu Dai, Shujian Huang, and Jiajun Chen. 2019.
\newblock \href {https://doi.org/10.18653/v1/N19-1259} {Target-oriented opinion
  words extraction with target-fused neural sequence labeling}.
\newblock In \emph{Proceedings of the 2019 Conference of the North {A}merican
  Chapter of the Association for Computational Linguistics: Human Language
  Technologies, Volume 1 (Long and Short Papers)}, pages 2509--2518,
  Minneapolis, Minnesota. Association for Computational Linguistics.

\bibitem[{He et~al.(2020)He, Liu, Gao, and Chen}]{he2020deberta}
Pengcheng He, Xiaodong Liu, Jianfeng Gao, and Weizhu Chen. 2020.
\newblock Deberta: Decoding-enhanced bert with disentangled attention.
\newblock \emph{arXiv preprint arXiv:2006.03654}.

\bibitem[{Huang et~al.(2021)Huang, Wang, Li, Liu, Zhang, Cheng, Yin, and
  Wang}]{First_Target_and_Opinion_then_Polarity}
Lianzhe Huang, Peiyi Wang, Sujian Li, Tianyu Liu, Xiaodong Zhang, Zhicong
  Cheng, Dawei Yin, and Houfeng Wang. 2021.
\newblock \href {http://arxiv.org/abs/arXiv:2102.08549} {First target and
  opinion then polarity: Enhancing target-opinion correlation for aspect
  sentiment triplet extraction}.

\bibitem[{Klinger and Cimiano(2014)}]{klinger-cimiano-2014-usage}
Roman Klinger and Philipp Cimiano. 2014.
\newblock \href
  {http://www.lrec-conf.org/proceedings/lrec2014/pdf/85_Paper.pdf} {The {USAGE}
  review corpus for fine grained multi lingual opinion analysis}.
\newblock In \emph{Proceedings of the Ninth International Conference on
  Language Resources and Evaluation ({LREC}'14)}, pages 2211--2218, Reykjavik,
  Iceland. European Language Resources Association (ELRA).

\bibitem[{Koco{\'n} et~al.(2019{\natexlab{a}})Koco{\'n}, Mi{\l}kowski, and
  Za{\'s}ko-Zieli{\'n}ska}]{kocon-etal-2019-multi}
Jan Koco{\'n}, Piotr Mi{\l}kowski, and Monika Za{\'s}ko-Zieli{\'n}ska.
  2019{\natexlab{a}}.
\newblock \href {https://doi.org/10.18653/v1/K19-1092} {Multi-level sentiment
  analysis of {P}ol{E}mo 2.0: Extended corpus of multi-domain consumer
  reviews}.
\newblock In \emph{Proceedings of the 23rd Conference on Computational Natural
  Language Learning (CoNLL)}, pages 980--991, Hong Kong, China. Association for
  Computational Linguistics.

\bibitem[{Koco{\'n} et~al.(2021)Koco{\'n}, Radom, Kaczmarz-Wawryk, Wabnic,
  Zaj{\c a}czkowska, and Za{\'s}ko-Zieli{\'n}ska}]{11321849}
Jan Koco{\'n}, Jarema Radom, Ewa Kaczmarz-Wawryk, Kamil Wabnic, Ada Zaj{\c
  a}czkowska, and Monika Za{\'s}ko-Zieli{\'n}ska. 2021.
\newblock \href {http://hdl.handle.net/11321/849} {{AspectEmo} 1.0:
  Multi-domain corpus of consumer reviews for aspect-based sentiment analysis}.
\newblock {CLARIN}-{PL} digital repository.

\bibitem[{Koco{\'n} et~al.(2019{\natexlab{b}})Koco{\'n}, Zaśko-Zielińska, and
  Miłkowski}]{Kocon2019}
Jan Koco{\'n}, Monika Zaśko-Zielińska, and Piotr Miłkowski.
  2019{\natexlab{b}}.
\newblock {Multi-level analysis and recognition of the text sentiment on the
  example of consumer opinions}.
\newblock In \emph{Proceedings of the International Conference Recent Advances
  in Natural Language Processing, RANLP 2019}.

\bibitem[{Lango(2019)}]{lango2019tackling}
Mateusz Lango. 2019.
\newblock Tackling the problem of class imbalance in multi-class sentiment
  classification: An experimental study.
\newblock \emph{Foundations of Computing and Decision Sciences},
  44(2):151--178.

\bibitem[{Li et~al.(2021)Li, Wang, Zhang, hua Zhong, Yin, and
  He}]{more-fine-grained}
Yuncong Li, Fang Wang, Wenjun Zhang, Sheng hua Zhong, Cunxiang Yin, and
  Yancheng He. 2021.
\newblock \href {http://arxiv.org/abs/2103.15255} {A more fine-grained
  aspect-sentiment-opinion triplet extraction task}.

\bibitem[{Liu et~al.(2022)Liu, Li, and Li}]{liu-etal-2022-robustly}
Shu Liu, Kaiwen Li, and Zuhe Li. 2022.
\newblock \href {https://doi.org/10.18653/v1/2022.naacl-main.20} {A robustly
  optimized {BMRC} for aspect sentiment triplet extraction}.
\newblock In \emph{Proceedings of the 2022 Conference of the North American
  Chapter of the Association for Computational Linguistics: Human Language
  Technologies}, pages 272--278, Seattle, United States. Association for
  Computational Linguistics.

\bibitem[{Naglik and Lango(2023)}]{naglik23}
Iwo Naglik and Mateusz Lango. 2023.
\newblock Exploiting phrase interrelations in span-level neural approaches
  for aspect sentiment triplet extraction.
\newblock In \emph{Advances in Knowledge Discovery and Data Mining}, pages
  222--233, Cham. Springer Nature Switzerland.

\bibitem[{Nakayama et~al.(2018)Nakayama, Kubo, Kamura, Taniguchi, and
  Liang}]{doccano}
Hiroki Nakayama, Takahiro Kubo, Junya Kamura, Yasufumi Taniguchi, and Xu~Liang.
  2018.
\newblock \href {https://github.com/doccano/doccano} {{doccano}: Text
  annotation tool for human}.
\newblock Software available from https://github.com/doccano/doccano.

\bibitem[{Peng et~al.(2020)Peng, Xu, Bing, Huang, Lu, and Si}]{peng2020knowing}
Haiyun Peng, Lu~Xu, Lidong Bing, Fei Huang, Wei Lu, and Luo Si. 2020.
\newblock Knowing what, how and why: A near complete solution for aspect-based
  sentiment analysis.
\newblock In \emph{Proceedings of the AAAI conference on artificial
  intelligence}, volume~34, pages 8600--8607.

\bibitem[{Pontiki et~al.(2016)Pontiki, Galanis, Papageorgiou, Androutsopoulos,
  Manandhar, AL-Smadi, Al-Ayyoub, Zhao, Qin, De~Clercq, Hoste, Apidianaki,
  Tannier, Loukachevitch, Kotelnikov, Bel, Jim{\'e}nez-Zafra, and
  Eryi{\u{g}}it}]{pontiki-etal-2016-semeval}
Maria Pontiki, Dimitris Galanis, Haris Papageorgiou, Ion Androutsopoulos,
  Suresh Manandhar, Mohammad AL-Smadi, Mahmoud Al-Ayyoub, Yanyan Zhao, Bing
  Qin, Orph{\'e}e De~Clercq, V{\'e}ronique Hoste, Marianna Apidianaki, Xavier
  Tannier, Natalia Loukachevitch, Evgeniy Kotelnikov, Nuria Bel,
  Salud~Mar{\'\i}a Jim{\'e}nez-Zafra, and G{\"u}l{\c{s}}en Eryi{\u{g}}it. 2016.
\newblock \href {https://doi.org/10.18653/v1/S16-1002} {{S}em{E}val-2016 task
  5: Aspect based sentiment analysis}.
\newblock In \emph{Proceedings of the 10th International Workshop on Semantic
  Evaluation ({S}em{E}val-2016)}, pages 19--30, San Diego, California.
  Association for Computational Linguistics.

\bibitem[{Pontiki et~al.(2015)Pontiki, Galanis, Papageorgiou, Manandhar, and
  Androutsopoulos}]{pontiki-etal-2015-semeval}
Maria Pontiki, Dimitris Galanis, Haris Papageorgiou, Suresh Manandhar, and Ion
  Androutsopoulos. 2015.
\newblock \href {https://doi.org/10.18653/v1/S15-2082} {{S}em{E}val-2015 task
  12: Aspect based sentiment analysis}.
\newblock In \emph{Proceedings of the 9th International Workshop on Semantic
  Evaluation ({S}em{E}val 2015)}, pages 486--495, Denver, Colorado. Association
  for Computational Linguistics.

\bibitem[{Pontiki et~al.(2014)Pontiki, Galanis, Pavlopoulos, Papageorgiou,
  Androutsopoulos, and Manandhar}]{pontiki-etal-2014-semeval}
Maria Pontiki, Dimitris Galanis, John Pavlopoulos, Harris Papageorgiou, Ion
  Androutsopoulos, and Suresh Manandhar. 2014.
\newblock \href {https://doi.org/10.3115/v1/S14-2004} {{S}em{E}val-2014 task 4:
  Aspect based sentiment analysis}.
\newblock In \emph{Proceedings of the 8th International Workshop on Semantic
  Evaluation ({S}em{E}val 2014)}, pages 27--35, Dublin, Ireland. Association
  for Computational Linguistics.

\bibitem[{Ptaszynski et~al.(2019)Ptaszynski, Pieciukiewicz, and
  Dyba{\l}a}]{ptaszynski2019results}
Michal Ptaszynski, Agata Pieciukiewicz, and Pawe{\l} Dyba{\l}a. 2019.
\newblock Results of the poleval 2019 shared task 6: First dataset and open
  shared task for automatic cyberbullying detection in polish twitter.
\newblock \emph{Proceedings of the PolEval 2019 Workshop}, page~89.

\bibitem[{Rybak et~al.(2020)Rybak, Mroczkowski, Tracz, and
  Gawlik}]{rybak-etal-2020-klej}
Piotr Rybak, Robert Mroczkowski, Janusz Tracz, and Ireneusz Gawlik. 2020.
\newblock \href {https://www.aclweb.org/anthology/2020.acl-main.111} {{KLEJ}:
  Comprehensive benchmark for polish language understanding}.
\newblock In \emph{Proceedings of the 58th Annual Meeting of the Association
  for Computational Linguistics}, pages 1191--1201, Online. Association for
  Computational Linguistics.

\bibitem[{Steinberger et~al.(2014)Steinberger, Brychc{\'\i}n, and
  Konkol}]{steinberger-etal-2014-aspect}
Josef Steinberger, Tom{\'a}{\v{s}} Brychc{\'\i}n, and Michal Konkol. 2014.
\newblock \href {https://doi.org/10.3115/v1/W14-2605} {Aspect-level sentiment
  analysis in {C}zech}.
\newblock In \emph{Proceedings of the 5th Workshop on Computational Approaches
  to Subjectivity, Sentiment and Social Media Analysis}, pages 24--30,
  Baltimore, Maryland. Association for Computational Linguistics.

\bibitem[{Troszy{\'n}ski and Wawer(2017)}]{troszynski2017czy}
Marek Troszy{\'n}ski and Aleksander Wawer. 2017.
\newblock Czy komputer rozpozna hejtera? wykorzystanie uczenia maszynowego (ml)
  w jako{\'s}ciowej analizie danych.
\newblock \emph{Przegl{\k{a}}d Socjologii Jako{\'s}ciowej}, 13(2):62--80.

\bibitem[{Wawer and Ogrodniczuk(2017)}]{wawer2017results}
Aleksander Wawer and Maciej Ogrodniczuk. 2017.
\newblock Results of the poleval 2017 competition: Sentiment analysis shared
  task.
\newblock In \emph{8th Language and Technology Conference: Human Language
  Technologies as a Challenge for Computer Science and Linguistics}.

\bibitem[{Wu et~al.(2020)Wu, Ying, Zhao, Fan, Dai, and Xia}]{wu-etal-2020-grid}
Zhen Wu, Chengcan Ying, Fei Zhao, Zhifang Fan, Xinyu Dai, and Rui Xia. 2020.
\newblock \href {https://doi.org/10.18653/v1/2020.findings-emnlp.234} {Grid
  tagging scheme for aspect-oriented fine-grained opinion extraction}.
\newblock In \emph{Findings of the Association for Computational Linguistics:
  EMNLP 2020}, pages 2576--2585, Online. Association for Computational
  Linguistics.

\bibitem[{Xu et~al.(2021)Xu, Chia, and Bing}]{span-level}
Lu~Xu, Yew~Ken Chia, and Lidong Bing. 2021.
\newblock Learning span-level interactions for aspect sentiment triplet
  extraction.
\newblock In \emph{Proceedings of the 59th Annual Meeting of the ACL and the
  11th IJCNLP (Volume 1: Long Papers)}, pages 4755--4766. Association for
  Computational Linguistics.

\bibitem[{Xu et~al.(2020)Xu, Li, Lu, and Bing}]{jet}
Lu~Xu, Hao Li, Wei Lu, and Lidong Bing. 2020.
\newblock Position-aware tagging for aspect sentiment triplet extraction.
\newblock In \emph{Proceedings of the 2020 Conference on Empirical Methods in
  Natural Language Processing (EMNLP)}, pages 2339--2349. Association for
  Computational Linguistics.

\bibitem[{Zhang et~al.(2021)Zhang, Li, Deng, Bing, and Lam}]{T5}
Wenxuan Zhang, Xin Li, Yang Deng, Lidong Bing, and Wai Lam. 2021.
\newblock Towards generative aspect-based sentiment analysis.
\newblock In \emph{Proceedings of the 59th Annual Meeting of the ACL and the
  11th IJCNLP (Volume 2: Short Papers)}, pages 504--510. Association for
  Computational Linguistics.

\end{thebibliography}

\section{Language Resource References}
\label{lr:ref}
\bibliographystylelanguageresource{lrec_natbib}
\bibliographylanguageresource{languageresource}

\appendix
%\section{Appendix: How to Produce the \texttt{.pdf}}
%\label{sec:append-how-prod}
%In order to generate a PDF file out of the LaTeX file herein, when citing language resources, the following steps need to be performed:

% \begin{enumerate}
% \item \texttt{xelatex your\ paper\ name.tex}
% \item \texttt{bibtex your\ paper\ name.aux}
% \item \texttt{bibtex languageresource.aux}    *NEW*
% \item \texttt{xelatex your\ paper\ name.tex}
% \item \texttt{xelatex your\ paper\ name.tex}
% \end{enumerate}

\end{document}